\documentclass[10pt,twocolumn,letterpaper]{article}

\usepackage{cvpr}
\usepackage{times}
\usepackage{epsfig}
\usepackage{graphicx}
\usepackage{amsmath}
\usepackage{amssymb}
\usepackage{dsfont}
\usepackage{siunitx}
\usepackage{subfigure}
\usepackage{multirow}
\usepackage{caption}
\usepackage{arydshln}
\usepackage{url}            %
\usepackage{booktabs}       %
\usepackage{amsfonts}       %
\usepackage{nicefrac}       %
\usepackage{color}
\usepackage{multirow}

\usepackage[pagebackref=true,breaklinks=true,letterpaper=true,colorlinks,bookmarks=false]{hyperref}

 \makeatletter
 \DeclareRobustCommand\onedot{\futurelet\@let@token\@onedot}
 \def\@onedot{\ifx\@let@token.\else.\null\fi\xspace}
 \def\eg{e.g\onedot} \def\Eg{E.g\onedot}
 \def\ie{i.e\onedot}

 \makeatother

\DeclareRobustCommand{\figref}[1]{Figure~\ref{#1}}

\DeclareRobustCommand{\Figref}[1]{Figure~\ref{#1}}

\DeclareRobustCommand{\secref}[1]{Section~\ref{#1}}

\DeclareRobustCommand{\tableref}[1]{Table~\ref{#1}}

\newcommand{\myparagraph}[1]{\noindent \textbf{\emph{#1}}}

\newcommand{\invisible}[1]{}%

\graphicspath{{./fig/}{./fig/plots/}}

\cvprfinalcopy %

\def\cvprPaperID{2052} %

\begin{document}

\title{Generating Descriptions with Grounded and Co-Referenced People} %

\newcommand{\authSpace}{\ \ \ \ \ \ }
 \author{
 Anna Rohrbach$^{1}$ \authSpace Marcus Rohrbach$^{2}$ \authSpace Siyu  Tang$^{1,3}$ \authSpace  Seong Joon Oh$^{1}$ \authSpace Bernt Schiele$^{1}$\\
$^{1}$Max Planck Institute for Informatics, Saarland Informatics Campus, Germany\\
$^{2}$UC Berkeley, CA, United States \ \ $^{3}$Max Planck Institute for Intelligent Systems, T\"{u}bingen, Germany\\
}

\maketitle

\begin{abstract}
Learning how to generate descriptions of images or videos received major interest both in the Computer Vision and Natural Language Processing communities. While a few works have proposed to learn a grounding during the generation process in an unsupervised way (via an attention mechanism), it remains unclear how good the quality of the grounding is and whether it benefits the description quality. In this work we propose a movie description model which learns to generate description and jointly ground (localize) the mentioned characters as well as do visual co-reference resolution between pairs of consecutive sentences/clips. We also propose to use weak localization supervision through character mentions provided in movie descriptions to learn the character grounding. At training time, we first learn how to localize characters by relating their visual appearance to mentions in the descriptions via a semi-supervised approach. 
We then provide this (noisy) supervision into our description model which greatly improves its performance. Our proposed description model improves over prior work w.r.t.~generated description quality and additionally provides grounding and local co-reference resolution. We evaluate it on the MPII Movie Description dataset using automatic and human evaluation measures and using our newly collected grounding and co-reference data for characters.
\end{abstract}

\section{Introduction}

When humans talk about what they see, they not only use common objects and terms, but typically refer to reappearing entities, most commonly using names (``John'') and referential words such as pronouns (``he'', ``it''). To correctly generate descriptions with reappearing entities, one needs to understand and link them across sentences and visual appearances (images/frames).
Current image/video captioning datasets essentially ignore this aspect as they ask to independently describe each image/clip with a single sentence. At the same time, \eg visual storytelling \cite{huang16naacl} and movie description \cite{rohrbach16ijcv} ultimately require solving this problem. However, the first approaches on visual storytelling \cite{huang16naacl} so far have not taken it into account, and current movie description challenges and approaches \cite{rohrbach15cvpr,torabi15arxiv} abstract from it by looking at a single clip at a time and replacing all the character mentions with \eg ``Someone''.

In this work we address grounded co-reference resolution, with application to movie description. The most prominent entities in movies are the people or \emph{characters}. In fact, there is a long line of work which aims to link character mentions in movie or TV scripts with their visual tracks \cite{cour09cvpr,Everingham06bmvc,sivic09cvpr,tapaswi12cvpr,parkhi15,bojanowski13iccv,ramanathan14eccv}. %
However, all these works are already given the description for all movies where they want to predict the linking. In contrast we want to generate a description, while jointly linking it with the currently and previously depicted character's visual presence.
\begin{figure}[t]
\small 
\vspace{-0.4cm}
\includegraphics[width=\linewidth]{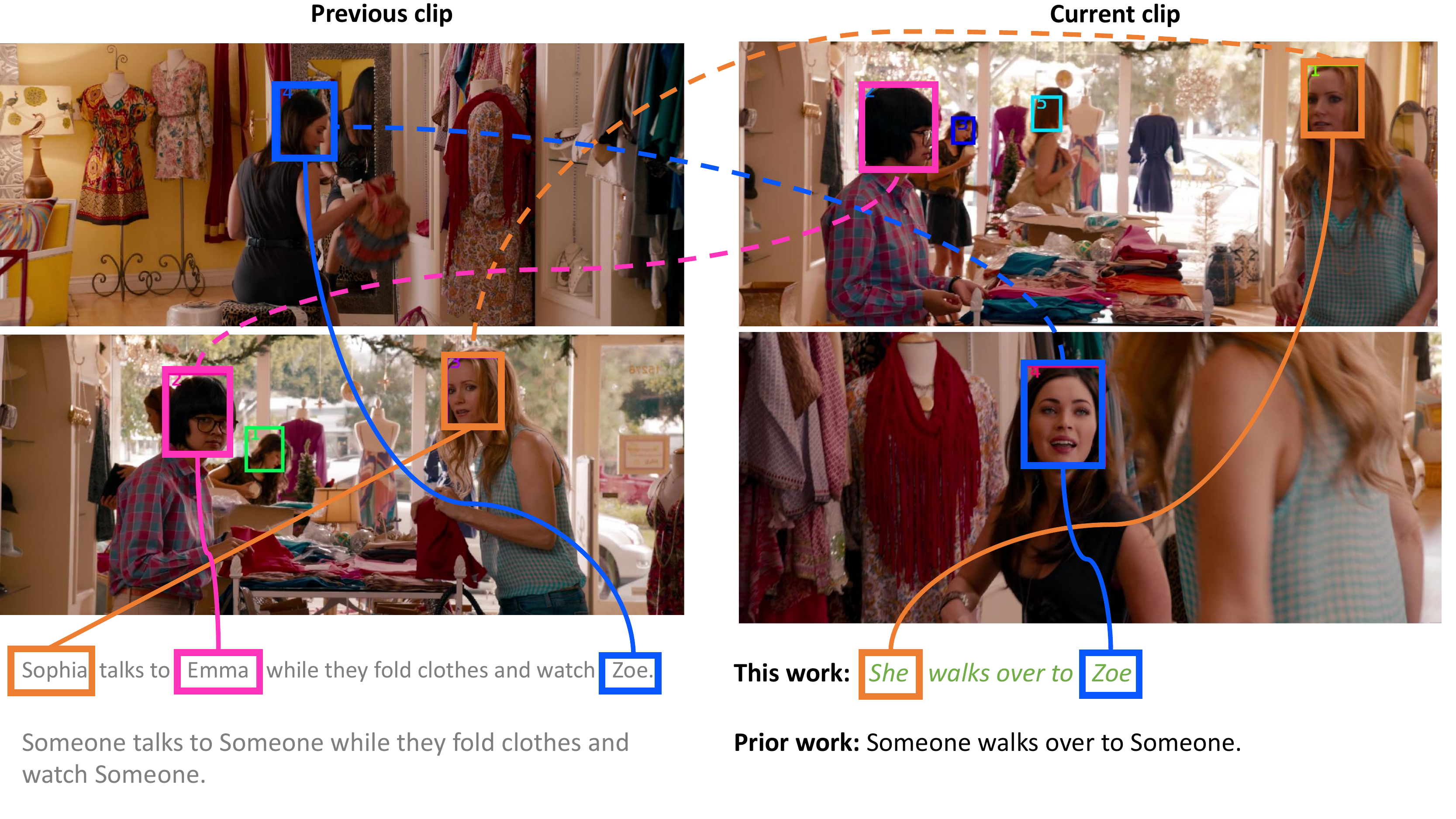} 
\vspace{-0.7cm}
\caption{\small Bring in the color: our task is to generate grounded and co-referenced descriptions %
for the current clip using pronouns and new or reappearing character IDs, which are grounded, \ie localized in the current clip (boxes and lines) and visually co-referenced to the previous clip (dashed lines). The visual grounding allows for co-reference to the previous clip/sentence which enables us using the pronoun ``she'' to refer to the first ID (Sophia).
}
\vspace{-0.5cm}
\label{cvpr17:fig:teaser}
\end{figure}
Specifically, the task we address in this work is to \emph{generate descriptions} for movies and at the same time localize or \emph{ground} the characters, recognize their gender and refer to them consistently, \ie \emph{co-reference} them across sentences, as visualized in \Figref{cvpr17:fig:teaser}.
Importantly, rather than trying to obtain consistent ids in the entire movie, we focus on robust \emph{local} co-reference resolution on \emph{two consecutive sentences/clips}. 
We argue that local co-reference resolution is an important problem on itself. On the one hand there are many characters without proper names and/or with only a few occurrences, which can and should be resolved locally, \eg ``\emph{The priest} takes their vows. \emph{He} declares them wife and husband''. On the other hand, there are many hard decisions which have to be made locally, \eg which character to describe and whether a character should be referenced by proper name or pronoun. %
To clarify, we do not generate the true proper names of the characters, but only identities with gender. We use a predefined set of names in our examples (\eg Sophia). In future work we believe the true names could be extracted either from dialog, or from one/a few annotations per character.
Approaching the joint description and grounding task requires three main ingredients: we need to \emph{localize the characters}, we need to decide which character(s) to \emph{pay attention to}, and we need to \emph{co-reference} visual characters' appearances in neighboring sentences/clips.
In \secref{cvpr17:sec:characterdetectionandtracking} we detail how we approach \emph{character localization} using head detection and tracking via a two-stage clustering approach. %
While generating the sentence, we advocate to \emph{jointly} decide which character to \emph{pay attention to} and if and how to \emph{co-reference} it to the previous grounded characters. In \secref{cvpr17:sec:generating}, we propose to adapt the attention mechanism \cite{bahdanau2014neural,xu15icml} for this and extend it to attend \emph{jointly} over both problems: grounding (\ie track selection) and co-reference (\ie track linking). A key insight is that this can not be learned purely from sentence supervision for generation. Instead, we supervise the joint-attention mechanism with automatically obtained linking of character mentions and tracks (\secref{cvpr17:sec:grounder}). We note that at test time this supervision is not available and the system has learned, how to jointly ground, co-reference, and describe.

The contributions of our paper include: a) a new task of movie description with grounded and co-referenced characters; to foster research in this direction we will share our newly collected co-reference annotations and grounding of character mentions in the MPII-MD dataset (\secref{cvpr17:sec:dataset}); %
b) a novel approach which addresses this problem by jointly learning to ground the described characters and perform local co-reference resolution between the neighboring clips; c) a robust automatic way of obtaining linking between character mentions in text and visual tracks in video, which we use to supervise our description approach and which we show is essential for the co-reference resolution task.

\section{Related Work}
Our work aims to do three tasks jointly: generating video descriptions, grounding, and co-reference resolution. We review related work in these three directions with a focus on works which attempt multiple tasks at once. As we focus on people grounding and co-reference, we also discuss the related work on person re-identification and track naming.

\myparagraph{Description generation.} 
Generating natural language about visual content has received large interest since the emergence of recurrent networks. Typically the focus is to generate a single sentence about a single image \cite{donahue15cvpr,karpathy15cvpr,mao15iclr,vinyals15cvpr,xu15icml}, video \cite{donahue15cvpr,guadarrama13iccv,rohrbach14gcpr,rohrbach13iccv,thomason14coling}, or most closely to this work, movie clip \cite{rohrbach15gcpr,venugopalan15iccv}. Several works also produce grounding while generating the description: \cite{xu15icml} propose an attention mechanism to ground each word to spatial CNN image features, \cite{you16cvpr} extend this to bounding boxes, \cite{yao2015iccv} to video frames, and \cite{zanfir16accv} to spatial-temporal proposals. \cite{liu17aaai} look into evaluating attention correctness for image captioning. \cite{johnson16cvpr} take a different direction and build a model which describes the entire image by jointly predicting large number of bounding boxes and a corresponding short phrase for each box. \cite{lin15bmvc} parse the visual 3D scene and generate coherent multi-sentence descriptions where the objects are grounded in 3D cuboids. Multi-sentence image/video description has also been explored in \eg  \cite{huang16naacl,rohrbach14gcpr,shin16icip,yu16cvpr}.

\myparagraph{Grounding objects in images/video.}
Grounding nouns as well as complex natural language expressions in images \cite{hu16cvpr,kong14cvpr,mao16cvpr,plummer15iccv,rohrbach16eccv,wang2016cvpr,yu16eccv} and video \cite{lin14cvpr,yu13acl} has recently received increased interest. The focus in our work is to localize people in a video while mentioning them in a generated sentence. For example, when mentioning a character who is jogging in a park, we want to localize this person in the video. Additionally we are interested in obtaining visual tracks for character mentions in text, for which rely on the semi-supervised grounding approach from \cite{rohrbach16eccv}.

\myparagraph{Co-reference resolution.}
Co-reference resolution is the task defined in linguistic community \cite{bergsma06acl}, where the goal is to establish correct links between named entities and references to them, \eg pronouns. \cite{ramanathan14eccv} address co-reference resolution in TV show descriptions with a bidirectional optimization using character visual appearance and linguistic co-reference resolution features.

\myparagraph{Person re-identification.}
Person re-identification from face/head images is a well studied problem and recently many deep learning based approaches have been proposed to address it \cite{Li2014CvprDeepReID,parkhi15bmvc,Schroff2015ArxivFaceNet,Sun2014ArxivDeepId2plus,Taigman2014CvprDeepFace,Zhou2015ArxivNaiveDeepFace}. Our work is related to this line of work as we aim to re-identify characters between two video clips while generating a video description.

\myparagraph{Linking tracks to names.}
Related works \cite{cour09cvpr,Everingham06bmvc,ramanathan14eccv,sivic09cvpr,tapaswi12cvpr} propose datasets for character identification targeting TV shows, which rely on alignment of video to TV scripts. The goal is to track faces in the video and assign names to them. Typically the tracks include background characters. \cite{bojanowski13iccv} attack the problem of learning a joint model of actors and actions in movies using weak supervision provided by scripts. \cite{parkhi15} propose a multiple instance learning based approach which focuses on recognizing background characters, and show significant improvement over prior work. There are two differences between ours and these prior works. First, we aim to re-identify characters locally, without ever seeing them before. Second, when obtaining the matching between names and tracks, our goal is to predict the grounding for a given character, not to name all the tracks.

\section{A Dataset for Grounded and Co-Referenced Characters}
\label{cvpr17:sec:dataset}
One of the goals in this work is to learn the visual co-reference resolution. To address and evaluate this task we collected annotations both on language and visual sides. On the language side we want to know when different mentions actually refer to the same person. On the visual side we require grounding of names to visual appearances. Towards these goals we collect new annotations for character co-reference resolution and grounding for the MPII Movie Description (MPII-MD) dataset \cite{rohrbach15cvpr}.

\myparagraph{Co-reference annotations for character mentions.}
\label{cvpr17:sec:dataset:coref}

\newcommand{\mpiicoref}{MPII-MD Co-ref+Gender}

In the  first step, we aim to label all the character mentions in the movie descriptions of the MPII-MD. The standard version of the descriptions consists of sentences with all character names replaced with ``Someone'' and multiple names (\eg ``Ann and Bob'') with ``people''.  Along with the transformed descriptions, the MPII-MD dataset provides the original descriptions with all the character names preserved. We rely on these and run the Stanford Named Entity Recognizer (NER) \cite{finkel05acl} and obtain our initial name list. We perform manual cleaning and filter out non-human related entities. We also manually check for names missed by NER and add them to our list. With the final name list we label the names in the entire dataset which includes many instances missed by the original NER pass. 
As the second step, we annotate names and co-references for each movie.
\Eg there might be different ways of referring to the same character (``Mary Jane'' as ``MJ''), so we link them together under one ``alias''. Additionally, we annotate the gender of all the characters. As the last step, we annotate pronouns ``he'' and ``she'' in all descriptions. When possible we link them to one of the existing names (with some exceptions for rare characters which were not named). In total we label {45,325} name mentions and {17,839} pronouns, see Table \ref{cvpr17:tbl:mentions_boxes_stat}. 
With this information we create our corpus \textbf{\mpiicoref}, where we transform the original MPII-MD descriptions so that every character mention, which appears in a previous sentence, is replaced with ``MaleCoref''/``FemaleCoref'', otherwise with ``MaleName''/``FemaleName''. We emphasize that this is the only difference to the standard MPII-MD, \ie the video clips and splits are identical.

\myparagraph{Grounded character annotations.}
\label{cvpr17:sec:grounding_annotations}
To evaluate the correctness of character grounding we annotate some characters with bounding boxes in video frames. For a subset of movies from MPII-MD Training, Validation and Test set we randomly select sentences and annotate all the mentioned characters. Specifically, whenever the character is mentioned in the sentence and is visible in the corresponding clip, we annotate a few frames of the clip with his/her head bounding boxes. As we also want to evaluate the co-reference correctness, we additionally annotate pairs of consecutive sentences/clips from the Test set. In total we label {2,649} bounding boxes with names, see Table \ref{cvpr17:tbl:mentions_boxes_stat}.

\newcommand{\midruleStat}{\cmidrule(rr){1-1} \cmidrule(rr){2-4} \cmidrule(rr){5-5}}
\begin{table}[t]
\center
\footnotesize
\begin{tabular}{lrrrr}
\toprule
           & Names & Pronouns & All Mentions & Boxes \\
\midruleStat
Training   & 37,432 & 15,093 & 52,525 & 489 \\
Validation & 3,440  & 1,092  & 4,532  & 412 \\
Test       & 4,453  & 1,654  & 6,107  & 1,748 \\
\midruleStat
Total       & 45,325  & 17,839  & 63,164  & 2,649 \\
\bottomrule
\end{tabular}
\caption{Left: number of annotated mentions, right: number of named bounding boxes, on MPII-MD \cite{rohrbach15cvpr}.}
\vspace{-0.3cm}
\label{cvpr17:tbl:mentions_boxes_stat}
\end{table}

\section{Visual Representations for Characters and their Context}
\label{cvpr17:sec:characterdetectionandtracking}

In this section our goal is to localize individual characters in video and extract visual representations informative of their appearance and context. Towards this goal we first detect, track, and extract localized representations for individual characters (\secref{cvpr17:sec:characters}), and then extract global representations which capture the scene and context not captured in localized representations (\secref{cvpr17:sec:holisticFeatures}).

\subsection{Character tracks and representations}
\label{cvpr17:sec:characters}
To localize the characters in movies we focus on localizing their heads as most of the time the head of a character is shown, but frequently not the full body. In contrast to prior work \cite{ramanathan14eccv} we do not only focus on frontal faces but also allow for more challenging views, \eg back view.
We detect the heads (\secref{cvpr17:sec:head-detection}) and track them with a two-step clustering approach, which is able to track across shot boundaries (\secref{cvpr17:sec:head-tracking}). We extract visual representations on the tracks, informative for estimating characters' identity, activity, gender, and importance  (\secref{cvpr17:sec:approach:trackrepresentations}).

\vspace{-0.3cm}
\subsubsection{Head detection}
\label{cvpr17:sec:head-detection}

We first detect all person instances in our videos using a head detector. Unlike conventional face detectors, our head detector can reliably detect profile faces and even back view heads. This is desirable because movies contain a large variety of view angles on heads. Our detector is based on the Faster R-CNN  \cite{girshick15iccv}. %
For training our head detector we collect head bounding box annotations over the PASCAL VOC 2010 trainval set. The dataset consists of 10,103 images of 7,372 head instances. 6,659 images do not have people, but we retain them as source of negatives. 
We make two modifications to the original Faster R-CNN configuration %
to make it more suitable for our head detection task. First, we account for small heads by adding smaller scale ``anchor boxes''. Anchor boxes refer to a default set of sliding window proposals from which Faster R-CNN regresses detection bounding boxes. Second, instead of doing hard negative mining by only considering proposals with ground truth overlap $>0$ and $\leq 0.5$ as negatives, we include any proposals with overlap $\leq 0.5$. This greatly improves the quality of our head detector by increasing the diversity of negative head training samples.
We run our detector %
on every frame of MPII-MD. We keep all the head detections with scores $\geq 0.5$ and both dimensions $\geq 40$ pixels.

\vspace{-0.3cm}
\subsubsection{Head tracking}
\label{cvpr17:sec:head-tracking}
After obtaining the head detections we aim to track them within the video clip. More specifically, we want to group all detections corresponding to the same person together. We need to take into account that the movies have shot boundaries (rapid changes in a camera viewpoint/angle). Thus the motion of a person can not be the only cue for tracking and we require an appearance cue to group together different views of the same character. This motivates our two-step approach, where we first group head detections within individual shots based on their motion and then further group the obtained tracks based on their appearance. 

To detect a shot boundary between two frames we rely on two features. First, we obtain color histograms on both frames and compute the Manhattan distance between the two. Second, we run the Kanade-Lucas-Tomasi (KLT) point tracker \cite{lucas81ai,tomasi91cmu}, initialized in the first frame with corner points from the minimum eigenvalue algorithm. We compute the ratio of points that are reliably tracked in the second frame. Based on these two characteristics we estimate the thresholds which allow us to detect shot boundaries and achieve high recall on a small set of manually annotated frame pairs w.r.t. to being a shot boundary. 
We select the parameters on a set of annotated frames and get the F-score 0.98. We try to detect all boundaries if possible and not produce too many false positives (wrong boundaries). Our tracking approach can deal with some false positives by clustering different tracks together based on appearance.

Our tracking framework is based on \cite{tang2015subgraph}, a multicut \cite{chopra1993,groetschel1989} tracker for pedestrians in street scene videos. The idea is to build a graph based on person detections in video, and then obtain the tracks by partitioning the graph into an optimal number of connected components, based on attractive and repulsive pairwise terms between pairs of detections. It is essentially a clustering based tracking formulation, which produces robust tracking results.
We adapt the multicut tracker to generate tracks for person heads in video clips. %
We cast our task as a two-level clustering problem. At the first level, we generate tracks from detections that are obtained on the consecutive frames within shots. 
To generate tracks from detections, we employ simple geometric features %
between detection bounding boxes. 
Given two bounding boxes $b$ and $b'$, where each has spatial-temporal location $(x, y, t)$ , scale $h$ and a corresponding image region $B$, we define the following variables: $\bar{h} = \frac{(h_b + h_{b'})}{2}$, $\Delta x = \frac{|x_b-x_{b'}|}{\bar{h}}, \Delta y = \frac{|y_b-y_{b'}|}{\bar{h}} ,\Delta h = \frac{|h_b-h_{b'}|}{\bar{h}},IOU = \frac{|B_d \cap B_{d'}|}{|B_d \cup B_{d'}|}$, where $IOU$ is the intersection over union of the two detection bounding boxes. The pairwise feature is defined as $(\Delta x, \Delta y, \Delta h,  IOU)$. Additionally, we add the quadratic terms of each feature to form a nonlinear mapping from feature space to the pairwise potentials. 

Next, we cluster the obtained tracks, selecting the ones that are at least 5 frames long for computational efficiency. For this we rely on the visual appearance features. For each track we mean pool the FaceVGG \cite{parkhi15bmvc} fc7 representations on the head crops. We then compute the $cosine$ distance between pairs of tracks and use $1 -$ distance as pairwise potentials in the second clustering step.

\vspace{-0.3cm}
\subsubsection{Track representations}
\label{cvpr17:sec:approach:trackrepresentations}
The representations extracted from the tracks should allow us to (re-)identify the characters, predict their activity and gender, and estimate if they should be described.

For re-identification of characters we again rely on the FaceVGG \cite{parkhi15bmvc} fc7 representation, referred to as $v^{head}$ in the following. We mean pool the track $t$ representation over all head crops clustered in this track and refer to it as $v^{head}(t)$. We discuss in \secref{cvpr17:sec:generating} how we estimate the similarity of two tracks for character re-identification in our pipeline.
We include the person body context which could be useful to \eg predict the person's activity. We extract the body region w.r.t. the head bounding box: 3 times wider and 6 times taller. We experiment with two visual features on the body region. First is a VGG \cite{simonyan15iclr} fc7 representation fine-tuned for 393 activities from the MPII human pose activity dataset \cite{pishchulin14gcpr}, provided by \cite{gkioxari15iccv}. We only use the body crop ignoring the additional context features as they would be similar across tracks and thus likely not help too much to distinguish tracks, but would significantly increase computation. Another feature we compute is ResNet \cite{he16cvpr} (pool5), trained on ImageNet \cite{deng09cvpr} for object classification. We mean pool both visual representations over all body crops in a track and refer to this as $v^{body}(t)$. In the experiments we specify if/which feature is being used.
We find, as also noted in \cite{parkhi15}, that the described characters are frequently in the front, center, and large compared to characters not described (background characters). Rather than manually defining a good function we provide the following track statistics $v^{stat}(t)$ and allow our approach to learn from this data: track length, mean and standard deviation of head width/height/center/detection score.

We do not extract designated gender features, as we find that $v^{head}$ and $v^{body}$ carry strong information about this aspect. It is straightforward to include even more targeted representation as part of future work. All the computed representations are normalized element-wise by first mean centering and then dividing by the standard deviation to improve learning subsequent functions with deep learning.

\subsection{Holistic video representations}
\label{cvpr17:sec:holisticFeatures}
In the previous section we discussed how and which localized features we extract for characters. To additionally capture context, objects, and scene information, important for movie description, we additionally rely on global representations provided by \cite{rohrbach15gcpr} for the MPII-MD dataset. We shortly review them in the following: 1) scores from 146 activity classifiers trained with Dense Trajectory features \cite{wang13iccv}; 2) scores from 99 object classifiers trained with LSDA \cite{hoffman14nips} responses; 3) scores from 18 scene classifiers trained with PLACES-CNN \cite{zhou14nips} responses. All the classifiers were trained in \cite{rohrbach15gcpr} using the words from descriptions as labels. The provided visual feature $v^{global}$ is a 263 dimensional concatenation of all three groups of scores.

\section{Generating Grounded and Co-Referenced Descriptions}
\label{cvpr17:sec:generating}
As discussed in the introduction, we focus on character grounding and local co-reference resolution, while generating the description. More specifically, we aim to predict the character grounding and do co-reference resolution given the previous sentence grounding. At test time this allows to \eg process the movie sequentially from start to end. In the following we rely on our transformed description corpus, \textbf{\mpiicoref}, described in \secref{cvpr17:sec:dataset:coref}.

\begin{figure*}[t]
\scriptsize
\center
\includegraphics[width=0.95\linewidth]{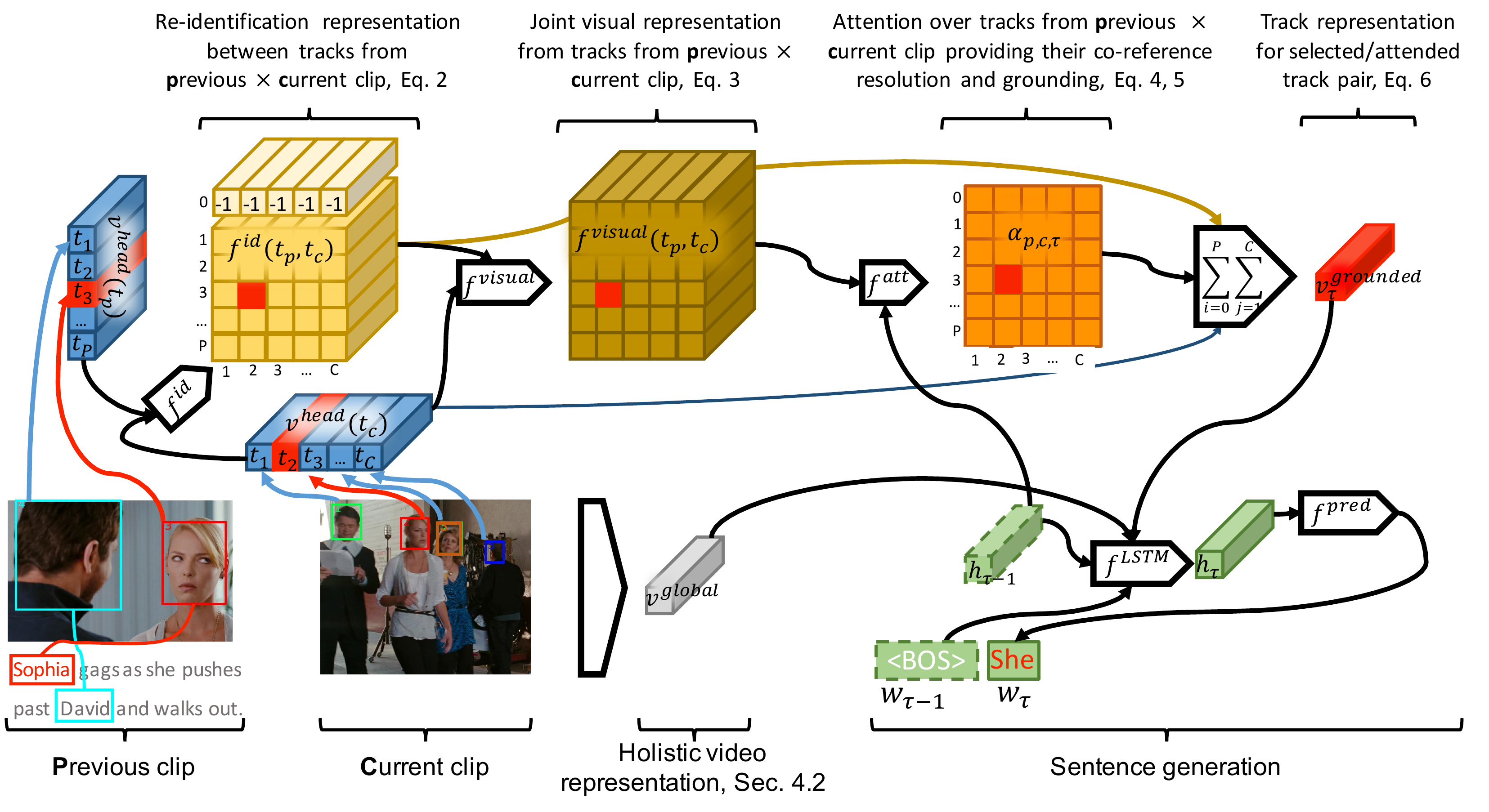} \vspace{-0.3cm}
\caption{Our model. Some components are omitted for clarity, \eg we omit the body and statistic representations.}
\label{cvpr17:fig:model}
\vspace{-0.3cm}
\end{figure*}

The key ideas of our approach are to predict grounding and co-reference resolution \emph{jointly} while generating the sentence (\secref{cvpr17:sec:generateAndGround}) and to learn grounding and co-reference with noisy supervision at training time obtained automatically by linking character mentions and tracks (\secref{cvpr17:sec:grounder}). \Figref{cvpr17:fig:model} provides an overview of our model.

\subsection{Predicting grounding and co-reference during sentence generation}
\label{cvpr17:sec:generateAndGround}
For generating sentences we rely on a recurrent LSTM \cite{hochreiter1997long} network as defined in \cite{zaremba2014learning}. To predict the hidden state at step $\tau$ of the sentence, we provide it with the previous word $w_{\tau-1}$ and hidden state $h_{\tau-1}$, as well as the current visual representation $v_{\tau}$:
 $h_{\tau} = f^{LSTM}([w_{\tau-1},v_{\tau}], h_{\tau-1})$
where $[,]$ denotes concatenation. The $f^{LSTM}$ has an additional hidden state or memory cell $c_t$ which is not exposed.
The word is then predicted as 
$w_{\tau} = f^{pred}( h_{\tau}) = Softmax(W^{pred} h_{\tau} + b^{pred}  )$
which can be supervised with the ground truth word $\hat{w}_{\tau}$.
Note that our vocabulary $w\in V$ does not contain any character names, but only  
$V^{person} = \{$\emph{MaleCoref,FemaleCoref,MaleName,FemaleName}$\}\subset V$.%

In the following we discuss how we obtain a $v_{\tau}$ which allows to predict the correct word and at the same time solve the grounding and co-reference problem. 
We formulate the problem in terms of tracks which are the result of the head tracking in \secref{cvpr17:sec:head-tracking}. We have tracks $t_c\in T^c$  in the \textbf{c}urrent clip $(C=|T^c|)$, %
 and tracks $t_p \in T^p $ in the \textbf{p}revious clip $(P=|T^p|)$%
. We always assume the sentences in the previous clip are already grounded to  tracks and only consider those tracks which correspond to mentions of characters in the sentence.
Whenever we generate a word $w_{\tau}$ which refers to a person $w_{\tau}\in V^{person}$, the task is to also select which track $t_{\hat{c}}$ it corresponds to in the current clip  and which track $t_{\hat{p}}$ in the previous clip. %
To account for the case when the person was not mentioned in the previous sentence we include $t_0$ in $T^p$ which represents a \texttt{null} track, which has to be selected to indicate that we describe a new name. As we are only modeling two consecutive clips at a time, this means if $t_{\hat{p}}=t_0$ we want to generate \emph{MaleName} or \emph{FemaleName} and \emph{MaleCoref} or \emph{FemaleCoref} otherwise.

\vspace{-0.3cm}
\paragraph{Track re-identification for visual co-reference.}
To estimate similarity of two tracks $t_p$ and $t_c$ we learn a weighting after element-wise multiplication\footnote{A note to our notation: We use superscript for names of variables and functions and subscript for indexes. $W$ is consistently used to represent learned multiplicative weights and $b$ to represent additive bias weights.}:
\begin{equation}
v^{id}(t_p,t_c) = v^{head}(t_p) \odot v^{head}(t_c)
\end{equation}
\vspace{-0.3cm}
\begin{equation}
f^{id}(t_p,t_c) = W^{id} v^{id}(t_p,t_c) 
\end{equation}
For $p=0$, which indicates that no similar track exists, we set $v^{id}(t_0,t_c)=-1$. 
In preliminary experiments we found that this works better than $0$, as values $v^{id}$ are close to $0$.

\vspace{-0.3cm}
\paragraph{Learning grounding and co-reference jointly.}
The goal of our approach is to select a track $t_{\hat{c}}$ and the corresponding previous track $t_{\hat{p}}$ which matches the person we are describing with the current word at time  $\tau$, in other words we ground this person in $t_{\hat{c}}$ and link it to $t_{\hat{p}}$. As noted above if $t_{\hat{p}}=t_0$   there is no previous track with the same identity as $t_{\hat{c}}$. 
We propose to jointly predict $\hat{c}$ and $\hat{p}$ using an attention mechanism which takes into account the re-identification and visual representations as well as the  hidden state $h_{\tau-1}$ of the recurrent LSTM network generating the description. 

The visual features are jointly embedded in the same space as the embedding learned for the hidden state:
\begin{multline}
f^{visual}(t_p,t_c) 
= W^{head}v^{head}(t_c)
+ W^{body}v^{body}(t_c)\\
+ W^{stat}v^{stat}(t_c) 
+ f^{id}(t_p,t_c) + b^v
\end{multline}
Afterwards visual and hidden state representation are element-wise multiplied and we learn a function to predict the attention $\alpha$. This is inspired by \cite{xu15icml}, who combine convolutional visual features and the recurrent hidden state in the same way to predict spatial attention. Conceptually different, we predict two aspects jointly, the grounding $t_p$ and linking $t_c$ of tracks  from different clips. 
\begin{multline}
\bar{\alpha}_{p,c,\tau} = f^{att}(t_p,t_c,\tau)=\\ W^{\alpha}\phi(W^h h_{\tau-1} + b^h) \odot \phi(f^{visual}(t_p,t_c) )+b^\alpha
\end{multline}
with the $htan$ non-linearity  $\phi(x) = \frac{e^x-e^{-x}}{e^x+e^{-x}}$.
The attention is normalized with softmax and then we use the predicted $\alpha$ in a weighted sum to get the new local visual representation:

\begin{equation}
\label{cvpr17:eq:attention}
\alpha_{p,c,\tau} = \frac{\exp(\bar{\alpha}_{p,c,\tau})}{\sum_{i=0}^P\sum_{j=1}^C{\exp(\bar{\alpha}_{i,k,\tau})}}
\end{equation}
\vspace{-0.3cm}
\begin{multline}
v^{grounded}_\tau=\sum_{i=0}^P\sum_{j=1}^C \alpha_{p,c,\tau}[v^{head}(t_c),\\
v^{body}(t_c),v^{stat}(t_c),v^{id}(t_p,t_c)].
\end{multline}
We use this together with the global/holistic video representation $v^{global}$ (see \secref{cvpr17:sec:holisticFeatures}) and the previous word  $w_{\tau-1}$ to predict the next hidden state of the recurrent LSTM network as discussed above:
$h_{\tau} = f^{LSTM}([v^{grounded},v^{global},w_{\tau-1}],h_{\tau-1})$.
\vspace{-0.3cm}
\paragraph{Supervising grounding and co-reference.}
While this system can be trained by only providing reference sentences as supervision, it is difficult to jointly correctly learn the grounding and co-reference resolution. We thus discuss in the next section how to obtain supervision for $\alpha_{p,c,\tau}$. Instead of annotating all characters mentions with tracks, we try to automatically predict the correct track $t$ for each character mention $w_{\tau}$ in the sentence. As we have ground truth co-reference on the text side for the entire training data (\secref{cvpr17:sec:dataset:coref}), we can construct the joint ground truth $\hat{\alpha}_{p,c,\tau}$ from the groundings per clip $\hat{\alpha}_{p,\tau}$, $\hat{\alpha}_{c,\tau}$. For all non-character words $w_{\tau}\notin V^{person}$, no supervision and thus no loss is provided. The losses from sentence supervision and grounding/co-reference supervision are weighted equally.

\subsection{Obtaining automatic supervision: \\linking character mentions and tracks}
\label{cvpr17:sec:grounder}
In this section we discuss how to ground or link character mention with id $m_\tau$ in text at position $\tau$ to a corresponding visual track $t_c$ in the video to provide ground truth $\hat{\alpha}_{c,\tau}$ used above. In contrast to sentence generation, here we explicitly use the character mentions $m$ (\eg "Harry") which appear in the text.
In other words we want to robustly choose the correct track for all character mentions. Note, that this is a slightly different task than in \eg \cite{parkhi15}, who aim to link all the visual tracks to correct names. %
To link the name mentions in text to tracks  we adapt the recently proposed semi-supervised approach GroundeR \cite{rohrbach16eccv}. This approach was initially proposed for the task of localizing text phrases within an image without localization supervision, \ie where the phrase is located. The main idea is to learn to  \emph{attend to} the right bounding box out of a set of proposals, by trying to reconstruct the phrase. We adapt this to our scenario by learning to localize a character $m_{\tau,k}$ in the set of tracks $T_k$ from clip $k$, where character $m$ is mentioned in the sentence $k$ at position $\tau$. We represent tracks with $v^{head}(t_{c,k})$ and encode character names $m$ together with an identifier of the $gender(m)\in \{M,F\}$ as separate word in an LSTM. Adding the gender allows the model to  exploit correlations with different visual appearance of male versus female people and thus helps selecting the right track.  In the special case when the sentence $k$ only contains a single name and the clip $k$ contains a single track, \ie $|T_k|=1$, we assume that grounding is correct and this information is used as additional supervision, thus enabling the semi-supervised setting of \cite{rohrbach16eccv}. To train the model we use pairs $([gender(m_{\tau,k}),m_{\tau,k}], \{v^{head}(t_{c,k})\}_{c\in\{1..C\}})$ and predict the grounding as the track with maximum attention from all the tracks in the clip. %

\section{Evaluation}

We start with evaluating the quality of our person head detection and tracking. Then we look at the quality of automatic linking between character names and tracks, obtained in \secref{cvpr17:sec:grounder}. Finally, we evaluate our complete pipeline for grounded movie description. We break down the evaluation in two parts: description quality and grounding quality.

\subsection{Head detection and tracking}
We evaluate our head detections and tracks on the collected bounding box annotations from \secref{cvpr17:sec:grounding_annotations}. Given the annotated bounding boxes we compute detection recall by looking whether there is a head detection in a given frame that has an Intersection Over Union (IOU) $\geq 0.5$ with the annotated head box. The track recall is computed similarly, based on the presence of the track that goes through the given frame while overlapping with the annotated box with IOU $\geq 0.5$. Table \ref{cvpr17:tbl:det_tracking}(left) shows recall on the Training, Validation and Test parts of the annotations.

\newcommand{\midruleDETGR}{ \cmidrule(rr){1-4} \cmidrule(rr){6-9}}

\begin{table}[t]
\center
\footnotesize
\begin{tabular}{@{\ }l@{\ \ }c@{\ \ \ }c@{\ \ }ccc@{\ \ }c@{\ \ }c@{\ \ }c@{\ \ }}
\midruleDETGR
Recall & Training & Val & Test & & Accuracy & Train & Val & Test  \\
\midruleDETGR
Detection & 82.00 & 65.78 & 84.73 & & GroundeR & 78.12 & 84.46 & 80.35 \\
Tracking  & 78.53 & 61.65 & 81.41 \\
\midruleDETGR
\end{tabular}
\vspace{-0.3cm}
\caption{(left) Detection and tracking recall on the annotated character heads. (right) GroundeR accuracy on the annotated names/bounding boxes (evaluated on the boxes covered by the tracks). In \%.}
\label{cvpr17:tbl:det_tracking}
\vspace{-0.4cm}
\end{table}

We analyze the missing recall of our head detector on the Training annotations. We find that there are multiple failure modes, such as motion blur, occlusion and head size (both small and large) contributing to the missing recall. On the well visible heads we achieve 93.2\% recall. %
The tracking recall is slightly lower than the detection recall, due to the short track rejection (see Section \ref{cvpr17:sec:head-tracking}). In particular, tracking can be hard when the head is observed from an unusual angle. Overall, we find that our annotations are rather challenging but the obtained performance is reasonable. We also note that our approach already works with just one good track for each character.%

\subsection{Linking characters to tracks}

For every clip we restrict the number of tracks to at most $50$. If more than $50$ tracks are available we sort them by length and keep the longest, otherwise we zero-complete the missing tracks. For the previous track we consider at most 7 candidate tracks in addition to the ``null'' track (no match among the previous tracks). Thus there are $8 \times 50$ possible choices to predict the character grounding and co-reference during sentence generation. We first train the GroundeR \cite{rohrbach16eccv} approach on Training movies only in order to  estimate the hyper parameters. Next we combine the Training, Validation and Test movies  and train GroundeR on this joint set. We evaluate the accuracy of the obtained predictions on the annotated pairs name/bounding box presented in \secref{cvpr17:sec:grounding_annotations}. For a given name we choose the top scoring track as the grounding prediction. For this track we then check whether it contains the annotated frame and overlaps with the annotated box by IOU $\geq 0.5$. Table \ref{cvpr17:tbl:det_tracking}(right) shows 
that GroundeR is able to quite robustly predict the correct track for a given character name.%

\subsection{Evaluating description quality}
\label{cvpr17:sec:eval:description}

We evaluate our approach in terms of description quality and compare it to a few baselines as well as prior work via an automatic as well as human evaluation. We report all the standard automatic measures in  \tableref{cvpr17:tbl:generation_automatic}. For human evaluation the human judges were provided with pairs of a reference sentence and a predicted sentence, and asked to compare them w.r.t. being helpful for a blind person to follow the events in the video \cite{rohrbach16ijcv}. The judges can decide that one sentence is better than the other or both are similar. Each pair is evaluated by three human judges. Afterwards for every system we compute the percentage of times when at least 2 out of 3 judges decided that the predicted sentence is similar or better than the reference. Table \ref{cvpr17:tbl:generation_automatic} presents the results of human evaluation in the last column.

The top part of the table contains the reference numbers from prior works on the standard version of the corpus. 
We cannot use attention supervision or evaluate grounding on standard MPII-MD, which are our core contributions. It is encouraging that our reduced model ``Our w/o $\alpha$'' achieves similar scores to prior work.   

The middle and bottom part of the table presents results on \mpiicoref, thus the numbers between the two settings are not directly comparable as the references changed which strongly affects the automatic evaluation measures. To address this we evaluate the approach Visual-Labels \cite{rohrbach15gcpr} on the transformed corpus. Unlike \cite{rohrbach15gcpr}, we do not ensemble multiple models. For a fair comparison with the Visual-Labels in the middle part of \tableref{cvpr17:tbl:generation_automatic}, we provide ablations that do not have access to the previous clip character grounding but instead select the 7 biggest previous tracks if sorted by track length multiplied by an average track area. 
We compare a variant of our approach without the body context features (``Our''), one with body features (``Our + Activity'') as described in \secref{cvpr17:sec:approach:trackrepresentations}, and one which removes the attention mechanism but uses the activity feature and encodes it jointly with the holistic feature (``Our + Activity w/o attention \& co-reference''). In the bottom part of \tableref{cvpr17:tbl:generation_automatic} we use the automatically obtained previous clip grounding (via \secref{cvpr17:sec:grounder}, which has access to the previous ground-truth sentence), so that different variants of our approach are comparable, as they obtain the same previous information. %
Here we compare ``Our'' and two variants of our approach with body features (``Our+Activity'', ``Our+ResNet''). We also ablate the impact of the grounding and co-reference supervision (``Our w/o $\hat{\alpha}$'') and the statistic features (``Our w/o statistic features''). 

\newcommand{\midruleGeneration}{\cmidrule(rr){1-1} \cmidrule(rr){2-5} \cmidrule(rr){6-6}}

\begin{table}[t]
\setlength{\tabcolsep}{3.4pt}
\footnotesize
\begin{tabular}{l@{}rrrrc}
\toprule
 & \multicolumn{4}{c}{Automatic} & {Human} \\
Approach & {Bleu-4} & {Metor} & {Rouge} & {CIDEr} & {judgment} \\
\midruleGeneration
\multicolumn{6}{c}{\bf{Standard MPII-MD with ``Someone''}} \\
Best of \cite{rohrbach15cvpr}       & 0.47 & 5.59 & 13.21 & 8.14 & -\\
Visual-Labels \cite{rohrbach15gcpr} & 0.80 & 7.03 & 16.02 & 9.98 & -\\
S2VT \cite{venugopalan15iccv}       & 0.64 & 7.10 & 15.69 & 6.96 & - \\
Our w/o $\hat{\alpha}$ & 0.84 & 6.43 & 16.10 & 10.66 & -\\
\midruleGeneration
\multicolumn{6}{c}{\bf{\mpiicoref}} \\
\multicolumn{6}{c}{\emph{without previous clip character grounding}}\\
Visual-Labels (no ensemble) &	0.66 & 5.21 & 13.94 & 10.34 & 11.8\\ %
Our + Act. w/o att.\&co-ref.   & 0.74 & 5.58 & 14.49 & 10.22 & 11.0\\ %
Our  & 0.67 & 5.06 & 13.17 & 10.89 & 14.8\\ %
Our + Activity & 0.71 & 5.31& 14.14& 11.33 & 15.0\\ %
\cmidrule(rr){1-1} \cmidrule(rr){2-5} \cmidrule(rr){6-6}
\multicolumn{6}{c}{\emph{with previous clip character grounding}}\\
Our w/o $\hat{\alpha}$ & 0.66 & 5.82 & 14.29 & 10.48 & 10.8\\ %
Our w/o statistic features & 0.75 & 5.81 & 14.97 & 11.65 & -\\
Our  &	0.68 & 5.81 & 15.33 & 11.70 & 14.0 \\ %
Our + Activity & 0.82 & 6.17 & 16.12 & 12.64 & 14.5\\ %
Our + ResNet & 0.88 & 6.00 & 15.70 & 11.76 & 13.0\\ %
\bottomrule
\end{tabular}
\vspace{-0.2cm}
\caption{Left: automatic / right: human evaluation of description generation on the test set of MPII-MD; for discussion see \secref{cvpr17:sec:eval:description}.}
\label{cvpr17:tbl:generation_automatic}
\vspace{-0.5cm}
\setlength{\tabcolsep}{6pt}
\end{table}
 
\begin{figure}[t]
\center
\includegraphics[width=\linewidth]{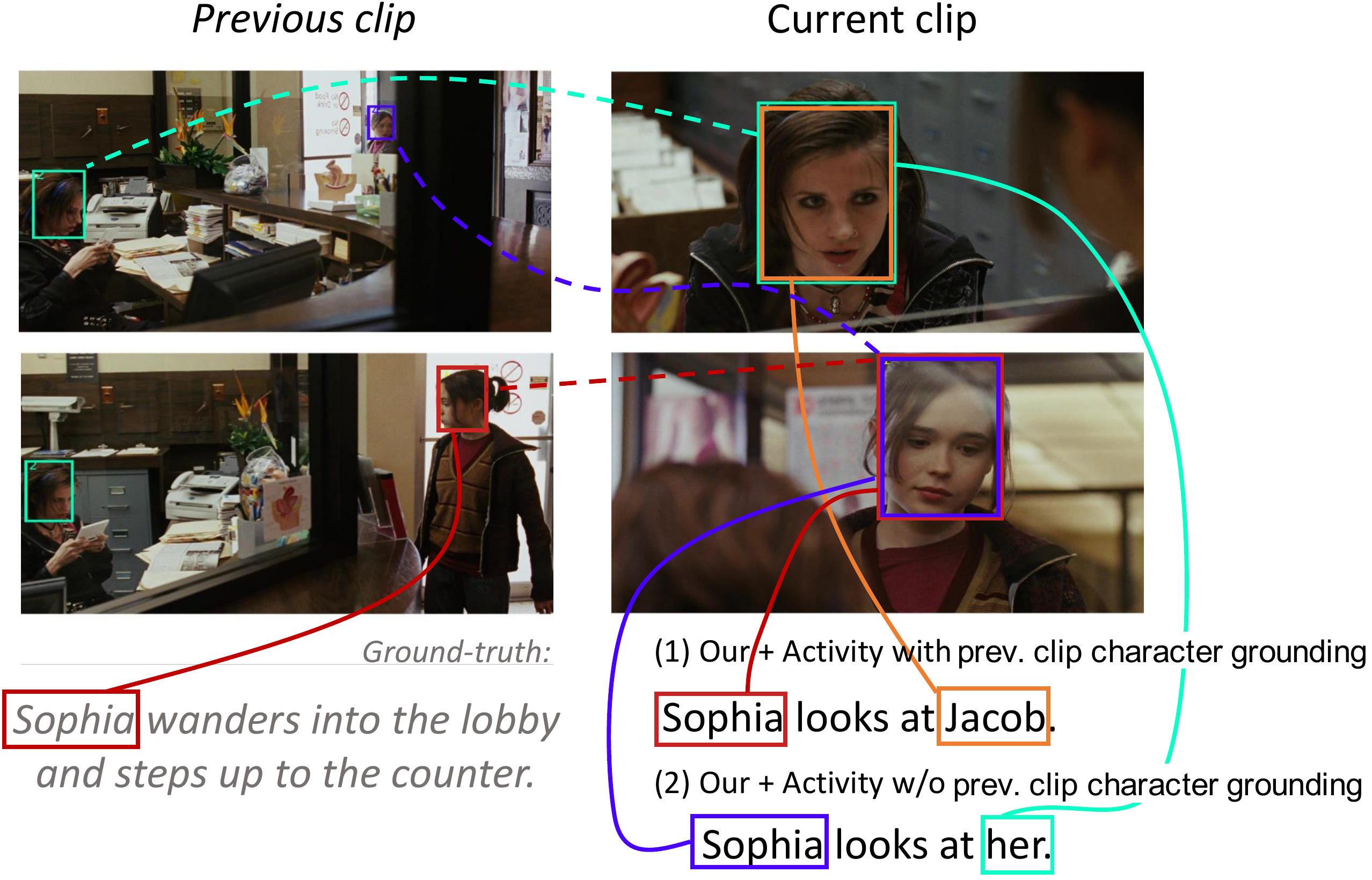} 
\caption{Supported by a visual co-reference to the previous clip, (2) correctly refers to a receptionist as \emph{`her'}, rather than \emph{`Jacob'}(1).}
\label{cvpr17:fig:qual_rebuttal}
\vspace{-0.4cm}
\end{figure}

From \tableref{cvpr17:tbl:generation_automatic} we see that: a) the systems ``Our'' / ``Our + Activity'' without previous clip character grounding achieve similar or better sentence quality than the Visual-Labels baseline; b) the variant with extra body context but without attention mechanism gets lower human score than our full system (11.0 vs. 15.0); c) providing grounding and co-reference supervision $\hat{\alpha}$ benefits the sentence quality; d) overall, body context features improve the scores, while the statistic features do not have a significant impact; e) the best result, according to human evaluation, is achieved by the variant of our approach ``Our + Activity'' \emph{without previous clip character grounding}. %
A possible explanation for this is as follows. In the automatically obtained previous clip's character grounding we might: a) link the characters to tracks correctly; b) link them incorrectly; c) miss some links if names are absent. In a) we follow the storyline of the movie. If we instead use the largest tracks of the previous clip, we bias the description of the current clip in a different way, \eg focus on the most salient characters. Thus, in some cases the obtained descriptions are ranked higher by the humans, as they only see the current clip in isolation (no story-line). In b), c) it is naturally more difficult to obtain a correct description of the current clip. See \figref{cvpr17:fig:qual_rebuttal} for an example.

\subsection{Evaluating grounding quality}
\label{cvpr17:sec:eval:grounding}

In this section we evaluate the correctness of the predicted grounding, co-reference and the generated character specific word $w_{\tau}\in \{$\emph{MaleCoref, FemaleCoref, MaleName, FemaleName}$\}$. We evaluate our predictions with respect to the manually obtained ground-truth (\secref{cvpr17:sec:grounding_annotations}) or automatically obtained ground-truth (\secref{cvpr17:sec:grounder}). For each of the named bounding boxes we obtain the track which overlaps with it most, for every character mention we obtain one or more associated ground-truth tracks. In total we obtain a set of $186$ sentences with manually obtained grounding and co-reference. For the automatic annotations we evaluate on a complete MPII-MD Test set ($6,578$ sentences).

We break down the evaluation in three parts: \emph{Grounding}, \emph{Grounding + Co-Reference}, \emph{Grounding + Co-Reference + $w_{\tau}$} (generated word). We compute precision and recall for each of these tasks and report the $F1$ score. Precision is computed as a percentage of predictions $\{{\alpha}_{p,c,\tau},w_{\tau}\}$, which are present in ground-truth. For the \emph{grounding} task we only check whether the track $t_c$ is present among ground-truth tracks. For \emph{co-reference} it has to be also correctly linked to the track $t_p$ from a previous clip. For the final task the predicted word $w_{\tau}$ with the track $t_c$ and predicted co-reference $t_p$ has to be present in the ground-truth. Recall is computed in a reversed way: for every ground-truth pair $\{\hat{\alpha}_{p,c,\tau},\hat{w_{\tau}}\}$ we check whether it is among the predictions.

\newcommand{\midruleground}{\cmidrule(rr){1-1} \cmidrule(rr){2-4} \cmidrule(rr){5-7}}
\begin{table}[t]
\begin{center}
\small
\begin{tabular}{@{}l@{\ }r@{\ }r@{\ \ }r@{\ }r@{\ }r@{\ \ }r@{}}
\toprule
\multicolumn{4}{r}{manual labeled subset} & \multicolumn{3}{@{}c@{}}{automatic gt, full set}\\
F1 score & Ground & +Co-Ref & +$w_{\tau}$ &Ground & +Co-Ref & +$w_{\tau}$ \\
\midruleground
\multicolumn{5}{l}{\textbf{Baselines with heuristic attention}}\\
\cite{rohrbach15gcpr} Center          & 59.21 & 19.33 & 13.83& 36.17 & 24.52 & 17.26\\
\cite{rohrbach15gcpr} LxA   & 69.58 & 23.93 & 18.80& 41.62 & 27.58 & 19.82\\
\cite{rohrbach15gcpr} LxA,Sim  & 69.58 & 39.05 & 6.07& 41.62 & 29.76 & 13.11\\
\midruleground
Our w/o $\hat{\alpha}$       & 64.60 & 21.75 & 13.47& 46.19 & 28.88 & 20.41\\
Our w/o stat.feat.           & 70.77 & 50.34 & 44.57& 46.34 & 38.14 & 32.87\\
Our                          & 69.17 & 53.92 & 49.55& 47.24 & 38.47 & 33.88\\
Our + Activity               & 71.99 & 50.54 & 45.63& 53.12 & 42.15 & 37.23\\
Our + ResNet                 & 69.76 & 51.51 & 46.54& 54.73 & 43.17 & 37.92\\
\midruleground
GroundeR gt         & 89.10 & 84.36 & 84.13\\
\bottomrule
\end{tabular}
\vspace{-0.3cm}
\caption{Grounding evaluation on test set. For discussion see \secref{cvpr17:sec:eval:grounding}.}
\vspace{-0.7cm}
\label{cvpr17:tbl:grounding_manual_eval}
\end{center}
\end{table}

The top part of \tableref{cvpr17:tbl:grounding_manual_eval} shows a set of baselines where we aim to obtain the grounding and co-reference resolution as a post-processing step after the sentence was generated. We use Visual-Labels \cite{rohrbach15gcpr} as a sentence generation baseline. We consider multiple heuristics to select the track: central position (Center), length times average area (LxA). Additionally we use a simple co-reference resolution method: if there are tracks in the previous clip, we pick the one which is most similar to the selected track as a co-reference (LxA,Sim). The similarity is estimated as $1 - cosine(v^{head}(t_c),v^{head}(t_p))$. The bottom part of the table lists the variants of our approach introduced earlier.
 
Table \ref{cvpr17:tbl:grounding_manual_eval}(left) presents the evaluation with the manually obtained ground-truth.  As we can see: a) the baselines are rather competitive in the grounding task, however they fall far below our approach in the co-reference task; b) grounding and co-reference supervision $\hat{\alpha}$ is very important to learn the co-reference prediction; c) statistics features, although they did not impact the description quality significantly, benefit the co-reference resolution; d) our approach is doing quite well in the final task, meaning that the language model correctly learns when to use co-references and recognizes the gender information.

\begin{figure}[t]
\small
\includegraphics[width=\linewidth]{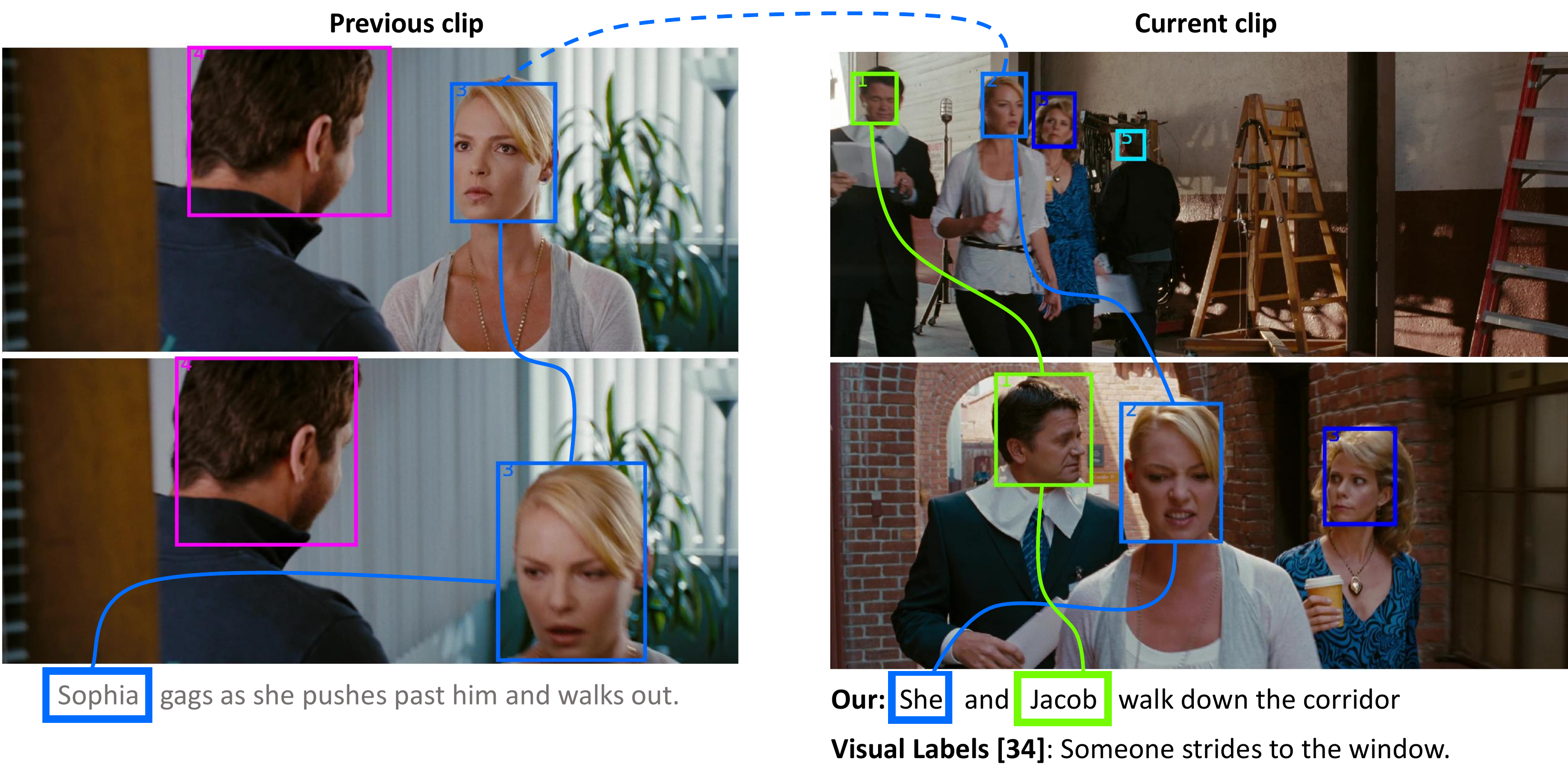}\vspace{0.2cm}
\includegraphics[width=\linewidth]{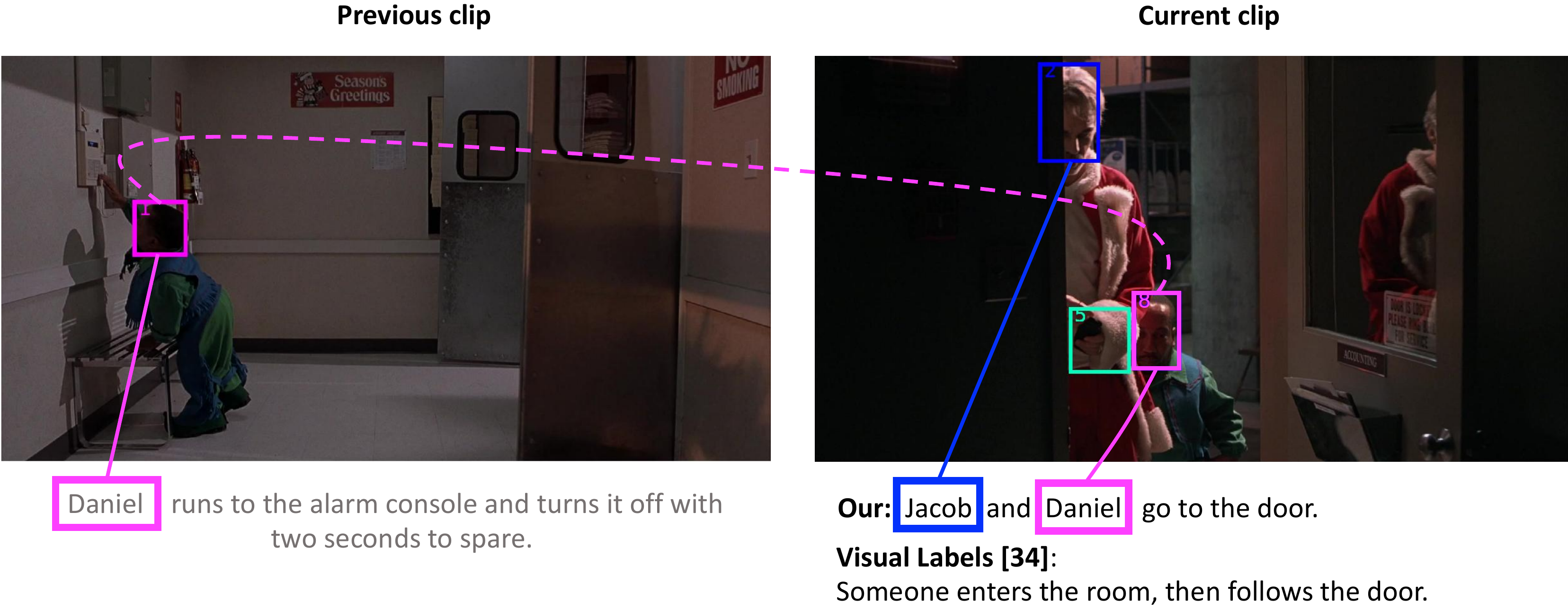} \vspace{-0.5cm}
\caption{Qualitative results of our approach on the grounded movie description task. Given a previous grounding we predict a sentence, grounding and co-reference.}
\label{cvpr17:fig:qual}
\vspace{-0.2cm}
\end{figure}

In the last line of Table \ref{cvpr17:tbl:grounding_manual_eval} we evaluate the quality of automatic ground-truth predictions from \secref{cvpr17:sec:grounder} with respect to our tasks. As we can see the predictions are overall quite reliable. Encouraged by that we perform the evaluation on this automatic ground-truth for the complete Test set, Table \ref{cvpr17:tbl:grounding_manual_eval}(right). We note, that the manually annotated set covers only 2.8\% of the full test set, so the results on the full test are more stable. We make the following observations: a) an ablation w/o statistic features again slightly drops in performance; b) all the baselines fall below our best approaches in all three tasks; this can be attributed to a more challenging data distribution: the complete test set contains sentences/clips where characters are absent and that has to be recognized correctly, while the manually annotated set always contains characters and is biased towards co-references; c) on this larger and more challenging test set we see that ``Our + Activity'' and ``Our + ResNet'' benefit from additional body features and achieve better performance than the basic variant ``Our''; one observation we make is that these two variants are more accurate with respect to presence/absence of people in the sentence/video which impacts the precision and thus the F1 score. In Figure \ref{cvpr17:fig:qual} we provide some qualitative examples with the predictions from our approach.

\section{Conclusions}

In this work we look at the novel task, generating descriptions with joint grounding and co-reference resolution of person mentions. %
We have proposed a novel approach, which relies on an attention mechanism that jointly learns to solve the grounding and co-reference resolution while learning to describe the video clip. %
Using an automatically learned linking between names and tracks we can provide supervision into our approach which significantly improves its ability to perform co-reference resolution. %
We demonstrate encouraging results in a complex task of grounded movie description and achieve improvements over multiple baselines. Our approach generates sentences of better quality than the baselines as shown by automatic and human evaluation. Overall, our approach can describe video, reason about persons identities, recognize their genders and  localize them in video.
We believe that this work is a first step towards fully coupling generation and grounding while performing image/video description. We will release the annotations and extracted tracks and hope that this will benefit other researchers who work on linguistic and/or visual co-reference resolution, movie question answering, visual storytelling, and multi-sentence video description.

\section*{Acknowledgments}
We would like to thank Trevor Darrell for helpful discussions. Marcus Rohrbach was supported by the Berkeley Artificial Intelligence Research (BAIR) Lab.

\small
\bibliographystyle{ieee}
\bibliography{biblioLong,rohrbach,related}
\end{document}